\pdfoutput=1

\documentclass[11pt]{article}

\usepackage{emnlp2021}

\usepackage{times}
\usepackage{latexsym}
\usepackage{booktabs}
\usepackage[T1]{fontenc}

\usepackage[utf8]{inputenc}

\usepackage{microtype}

\usepackage{graphicx}
\usepackage{amsmath}
\usepackage{amssymb}
\usepackage{amsthm}
\usepackage{booktabs}
\usepackage{algorithm}
\usepackage{algorithmic}
\usepackage{dsfont}
\usepackage{multirow}
\usepackage{bm}
\usepackage{pifont}
\usepackage{bbding}
\newcommand{\tabincell}[2]{\begin{tabular}{@{}#1@{}}#2\end{tabular}}

\newcommand\blfootnote[1]{%
  \begingroup
  \renewcommand\thefootnote{}\footnote{#1}%
  \addtocounter{footnote}{-1}%
  \endgroup
}

\title{Adaptive Proposal Generation Network for Temporal Sentence Localization in Videos}

\author{Daizong Liu\textsuperscript{1,2*}, Xiaoye Qu\textsuperscript{3*}, Jianfeng Dong\textsuperscript{4}, Pan Zhou\textsuperscript{1$\dagger$} \\
 \textsuperscript{1}The Hubei Engineering Research
Center on Big Data Security, School of \\ Cyber Science and Engineering, Huazhong University of Science and Technology \\
\textsuperscript{2}School of Electronic Information and Communication, \\ Huazhong University of Science and Technology \\
  \textsuperscript{3}Huawei Cloud \
  \textsuperscript{4}Zhejiang Gongshang University\\
  {\tt \normalsize \{dzliu,panzhou\}@hust.edu.cn, quxiaoye@huawei.com, dongjf24@gmail.com}}

\begin{document}
\maketitle
\begin{abstract}
We address the problem of temporal sentence localization in videos (TSLV). Traditional methods follow a \textit{top-down} framework which localizes the target segment with pre-defined segment proposals. Although they have achieved decent performance, the proposals are handcrafted and redundant. Recently, \textit{bottom-up} framework attracts increasing attention due to its superior efficiency. It directly predicts the probabilities for each frame as a boundary. However, the performance of \textit{bottom-up} model is inferior to the \textit{top-down} counterpart as it fails to exploit the segment-level interaction.
In this paper, we propose an \textbf{Adaptive} \textbf{Proposal} \textbf{Generation} \textbf{Network} (APGN) to maintain the segment-level interaction while speeding up the efficiency.
Specifically, we first perform a foreground-background classification upon the video and regress on the foreground frames to adaptively generate proposals. In this way, the handcrafted proposal design is discarded and the redundant proposals are decreased. Then, a proposal consolidation module is further developed to enhance the semantic of the generated proposals. Finally, we locate the target moments with these generated proposals following the \textit{top-down} framework. 
Extensive experiments on three challenging benchmarks show that our proposed APGN significantly outperforms previous state-of-the-art methods.
\vspace{-12pt}
\end{abstract}
\blfootnote{
\textsuperscript{$*$}Equal contributions. ~~~~\textsuperscript{$\dagger$}Corresponding author.}

\section{Introduction}
\begin{figure}[t!]
\centering
\includegraphics[width=0.48\textwidth]{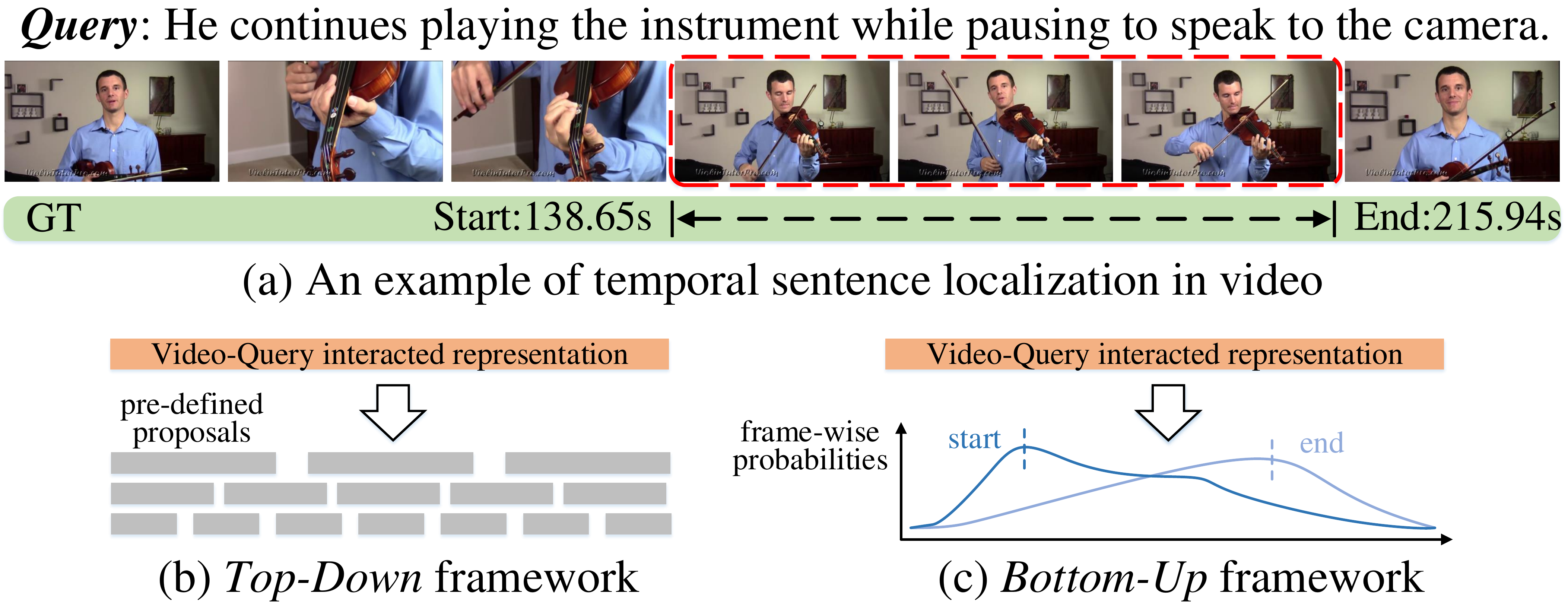}
\caption{(a) An example of temporal sentence localization in videos. (b) The \textit{Top-Down} framework predicts the confidence scores of a large number of pre-defined proposals for ranking. (c) The \textit{Bottom-Up} framework regresses the probabilities of all frames as start or end boundaries.}
\label{fig:introduction}
\vspace{-12pt}
\end{figure}

Temporal sentence localization in videos is an important yet challenging task in natural language processing, which has drawn increasing attention over the last few years due to its vast potential applications in 
information retrieval \cite{2019Dual,yang2020tree} and human-computer interaction \cite{singha2018dynamic}. It aims to ground the most relevant video segment according to a given sentence query. As shown in Figure \ref{fig:introduction} (a), most parts of video contents are irrelevant to the query (background) while only a short segment matches it (foreground). Therefore, video and query information need to be deeply incorporated to distinguish the fine-grained details of different video segments. 

Most previous works \cite{gao2017tall,chen2018temporally,zhang2019cross,yuan2019semantic,zhang2019learning,liu2021context,liu2020reasoning,liu2020jointly} follow the \textit{top-down} framework which pre-defines a large set of segment candidates (a.k.a proposals) in the video with sliding windows, and measures the similarity between the query and each candidate. The best segment is then selected according to the similarity. Although these methods achieve significant performance, they are sensitive to the proposal quality and present slow localization speed due to redundant proposals. Recently, several works \cite{rodriguez2020proposal,zhang2020span,yuan2019find} exploit the \textit{bottom-up} framework which directly predicts the probabilities of each frame as the start or end boundaries of segment. These methods are proposal-free and much more efficient. 
However, they neglect the rich information between start and end boundaries without capturing the segment-level interaction. Thus, the performance of \textit{bottom-up} models is behind the performance of \textit{top-down} counterpart thus far.

To avoid the inherent drawbacks of proposal design in the \textit{top-down} framework and maintain the localization performance, in this paper, we propose an adaptive proposal generation network (APGN) for an efficient and effective localization approach. 
Firstly, we perform boundary regression on the foreground frames to generate proposals, where foreground frames are obtained by a foreground-background classification on the entire video. In this way, the noisy responses on the background frames are attenuated, and the generated proposals are more adaptive and discriminative compared to the pre-defined ones.
Secondly, we perform proposal ranking to select target segment in a \textit{top-down} manner upon these generative proposals. As the number of proposals is much fewer than the pre-defined methods, the ranking stage is more efficient. 
Furthermore, we additionally consider the proposal-wise relations to distinguish their fine-grained semantic details before the proposal ranking stage.

To achieve the above framework, APGN first generates query-guided video representations after encoding video and query features and then predicts the foreground frames using a binary classification module. Subsequently, a regression module is utilized to generate a proposal on each foreground frame by regressing the distances from itself to start and end segment boundaries. After that, each generated proposal contains independent coarse semantic. 
To capture higher-level interactions among proposals, we encode proposal-wise features by incorporating both positional and semantic information, and represent these proposals as nodes to construct a proposal graph for reasoning correlations among them. Consequently, each updated proposal obtains more fine-grained details for following boundary refinement process. 

Our contributions are summarized as follows:
\vspace{-8pt}
\begin{itemize}
    \item We propose an adaptive proposal generation network (APGN) for TSLV task, which adaptively generates discriminative proposals without handcrafted design, thus making localization both effective and efficient.
    \vspace{-8pt}
    \item To further refine the semantics of the generated proposals, we introduce a proposal graph to consolidate proposal-wise features by reasoning their higher-order relations.
    \vspace{-8pt}
    \item We conduct experiments on three challenging datasets (ActivityNet Captions, TACoS, and Charades-STA), and results show that our proposed APGN significantly outperforms the existing state-of-the-art methods.
\end{itemize}

\section{Related Work}
Temporal sentence localization in videos is a new task introduced recently \cite{gao2017tall,anne2017localizing}, which aims to localize the most relevant video segment from a video with sentence descriptions. Various algorithms \cite{anne2017localizing,gao2017tall,chen2018temporally,zhang2019cross,yuan2019semantic,zhang2019learning,qu2020fine,Yang2021DVMR} have been proposed within the \textit{top-down} framework, which samples candidate segments from a video first, then integrates the sentence representation with those video segments individually and evaluates their matching relationships. Some of them \cite{anne2017localizing,gao2017tall} propose to use the sliding windows as proposals and then perform a comparison between each proposal and the input query in a joint multi-modal embedding space. To improve the quality of the proposals, \cite{zhang2019cross,yuan2019semantic} pre-cut the video on each frame by multiple pre-defined temporal scale, and directly integrate sentence information with fine-grained video clip for scoring. \cite{zhang2019learning} further build a 2D temporal map to construct all possible segment candidates by treating each frame as the start or end boundary, and match their semantics with the query information. Although these methods achieve great performance, they are severely limited by the heavy computation on proposal matching/ranking, and sensitive to the quality of pre-defined proposals.

\begin{figure*}[t!]
\vspace{-12pt}
\centering
\includegraphics[width=1.0\textwidth]{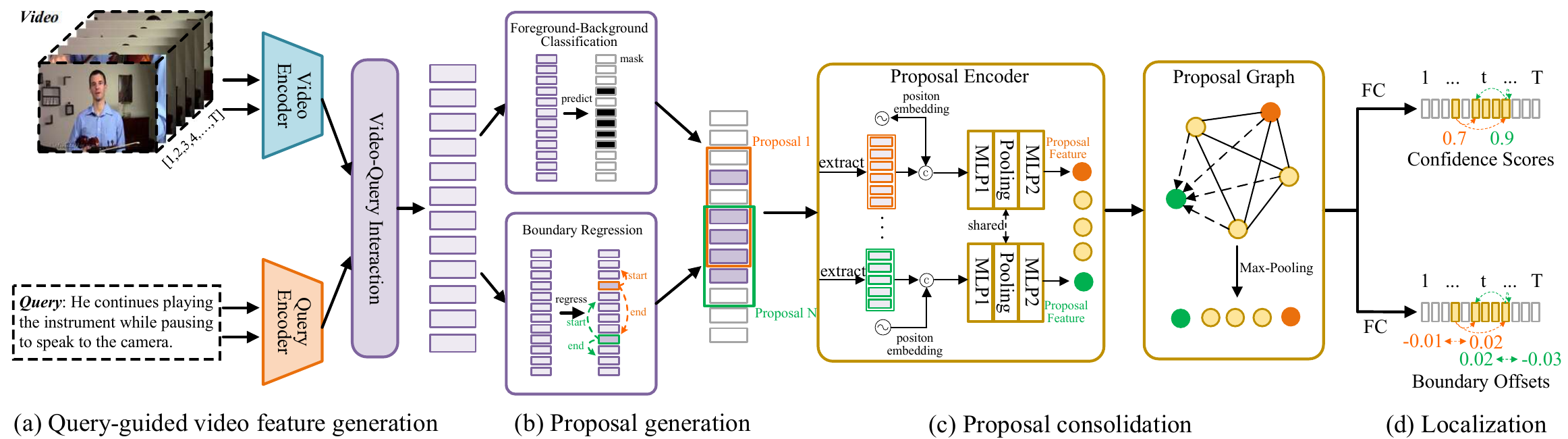}
\caption{Overall architecture of APGN. (a) Given a video and a query, we first encode and interact them to obtain query-guided video features. (b) Then, along with regressing boundaries on each frame, we perform foreground-background classification to identify the foreground frames whose corresponding predicted boundaries are further taken as the generated segment proposals. (c) We further encode each proposal and refine them using a graph convolutional network. (d) At last, we predict the confidence score and boundary offset for each proposal.}
\label{fig:pipeline}
\vspace{-8pt}
\end{figure*}

Recently, many methods \cite{rodriguez2020proposal,chenrethinking,yuan2019find,mun2020LGI,zeng2020dense,zhang2020span,nan2021interventional} propose to utilize the \textit{bottom-up} framework to overcome above drawbacks. They do not rely on the segment proposals and directly select the starting and ending frames by leveraging cross-modal interactions between video and query. Specifically, they predict two probabilities at each frame, which indicate whether this frame is a start or end frame of the ground truth video segment. Although these methods perform segment localization more efficiently, they lose the segment-level interaction, and the redundant regression on background frames may provide disturbing noise for boundary decision, leading to worse localization performance than \textit{top-down} methods. 

In this paper, we propose to preserve the segment-level interaction while speeding up the localization efficiency. Specifically, we design a binary classification module on the entire video to filter out the background responses, which helps model focus more on the discriminative frames. At the same time, we replace the pre-defined proposals with the generated ones and utilize a proposal graph for refinement. 

\section{The Proposed Method}
\subsection{Overview}
Given an untrimmed video $V$ and a sentence query $Q$, the TSLV task aims to localize the start and end timestamps $(\tau_s, \tau_e)$ of a specific video segment referring to the sentence query. 
We focus on addressing this task by adaptively generating proposals. 
To this end, we propose a binary classification module to filter out the redundant responses on background frames. Then, each foreground frame with its regressed start-end boundaries are taken as the generated segment proposal. In this way, the number of the generated proposals is much smaller than the number of pre-defined ones, making the model more efficient. Besides, a proposal graph is further developed to refine proposal features by learning their higher-level interactions. Finally, the confidence score and boundary offset are predicted for each proposal. Figure \ref{fig:pipeline} illustrates the overall architecture of our APGN. 

\subsection{Feature Encoders}
\noindent \textbf{Video encoder.} Given a video $V$, we represent it as $V=\{v_t\}_{t=1}^T$, where $v_t$ is the $t$-th frame and $T$ is the length of the entire video. We first extract the features by a pre-trained network, and then employ a self-attention \cite{vaswani2017attention} module to capture the long-range dependencies among video frames. We also utilize a Bi-GRU \cite{chung2014empirical} to learn the sequential characteristic. The final video features are denoted as $\bm{V}=\{\bm{v}_t\}_{t=1}^T \in \mathbb{R}^{T\times D}$, where $D$ is the feature dimension.

\noindent \textbf{Query encoder.} Given a query $Q=\{q_n\}_{n=1}^N$, where $q_n$ is the $n$-th word and $N$ is the length of the query. Following previous works \cite{zhang2019cross,zeng2020dense}, we first generate the word-level embeddings using Glove \cite{pennington2014glove}, and also employ a self-attention module and a Bi-GRU layer to further encode the query features as $\bm{Q}=\{\bm{q}_n\}_{n=1}^N \in \mathbb{R}^{N\times D}$.

\noindent \textbf{Video-Query interaction.} After obtaining the encoded features $\bm{V},\bm{Q}$, we utilize a co-attention mechanism \cite{lu-etal-2019-debug} to capture the cross-modal interactions between video and query features. Specifically, we first calculate the similarity scores between $\bm{V}$ and $\bm{Q}$ as:
\begin{equation}
    \bm{S} = \bm{V}(\bm{Q}\bm{W}_S)^{\text{T}} \in \mathbb{R}^{T\times N},
\end{equation}
where $\bm{W}_S \in \mathbb{R}^{D\times D}$ projects the query features into the same latent space as the video. Then, we compute two attention weights as:
\begin{equation}
    \bm{A} = \bm{S}_r (\bm{Q}\bm{W}_S) \in \mathbb{R}^{T\times D}, \bm{B} = \bm{S}_r \bm{S}_c^{\text{T}} \bm{V} \in \mathbb{R}^{T\times D},
\end{equation}
where $\bm{S}_r$ and $\bm{S}_c$ are the row- and column-wise softmax results of $\bm{S}$, respectively. We compose the final query-guided video representation by learning its sequential features as follows:
\begin{equation}
    \widetilde{\bm{V}} = \text{BiGRU}([\bm{V};\bm{A};\bm{V}\odot \bm{A};\bm{V}\odot \bm{B}]) \in \mathbb{R}^{T\times D},
\end{equation}
where $\widetilde{\bm{V}}=\{\widetilde{\bm{v}}_t\}^{T}_{t=1}$, $\text{BiGRU}(\cdot)$ denotes the BiGRU layers, $[;]$ is the concatenate operation, and $\odot$ is the element-wise multiplication.

\subsection{Proposal Generation}
Given the query-guided video features $\widetilde{\bm{V}}$, we aim to generate the proposal tuple $(t, l_{s}^{t}, l_{e}^{t})$ based on each foreground frame $v_t$, where $l_{s}^{t}, l_{e}^{t}$ denotes the distances from frame $v_t$ to the starting and ending segment boundaries, respectively. To this end, we first perform binary classification on the whole frames to distinguish the foreground and background frames, and then treat the foreground ones as positive samples and regress the segment boundaries on these frames as generated proposals.

\noindent \textbf{Foreground-Background classification.} In the TSLV task, most videos are more than two minutes long while the lengths of annotated target segments only range from several seconds to one minute (e.g. on ActivityNet Caption dataset). Therefore, there exists much noises from the background frames which may disturb the accurate segment localization. To alleviate it, we first classify the background frames and filter out their responses in latter regression. By distinguishing the foreground and background frames with annotations, we design a binary classification module with three full-connected (FC) layers to predict the class $y_t$ on each video frame. Considering the unbalanced foreground/background distribution, we formulate the balanced binary cross-entropy loss as:
\begin{equation}
\small{\mathcal{L}_{class}=-\sum_{t=1}^{T_{back}}\frac{T_{back}}{T}\text{log}(y_t)-\sum_{t=1}^{T_{fore}}\frac{T_{fore}}{T}\text{log}(1-y_t)},
\end{equation}
where $T_{fore},T_{back}$ are the numbers of foreground and background frames. $T$ is the number of total video frames. Therefore, we can differentiate between frames from foreground and background during both training and testing. 

\noindent \textbf{Boundary regression.} With the query-guided video representation $\widetilde{\bm{V}}$ and the predicted binary sequence of 0-1, we then design a boundary regression module to predict the distance from each foreground frame to the start (or end) frame of the video segment that corresponds to the query. We implement this module by three 1D convolution layers with two output channels. Given the predicted distance pair $(l_{s}^{t}, l_{e}^{t})$ and ground-truth distance $(g_{s}^{t}, g_{e}^{t})$, we define the regression loss as:
\begin{equation}
\small{\mathcal{L}_{reg}=\frac{1}{T_{fore}}\sum_{t=1}^{T_{fore}} (1-\text{IoU}((t, l_{s}^{t}, l_{e}^{t}), (t, g_{s}^{t}, g_{e}^{t})))},
\end{equation}
where $\text{IoU}(\cdot)$ computes the Intersection over Union (IoU) score between the predicted segment and its ground-truth. After that, we can represent the generated proposal as tuples $\{( t, l_{s}^{t}, l_{e}^{t})\}_{t=1}^{T_{fore}}$ based on the regression results of the foreground frames.

\begin{figure}[t!]
\centering
\includegraphics[width=0.48\textwidth]{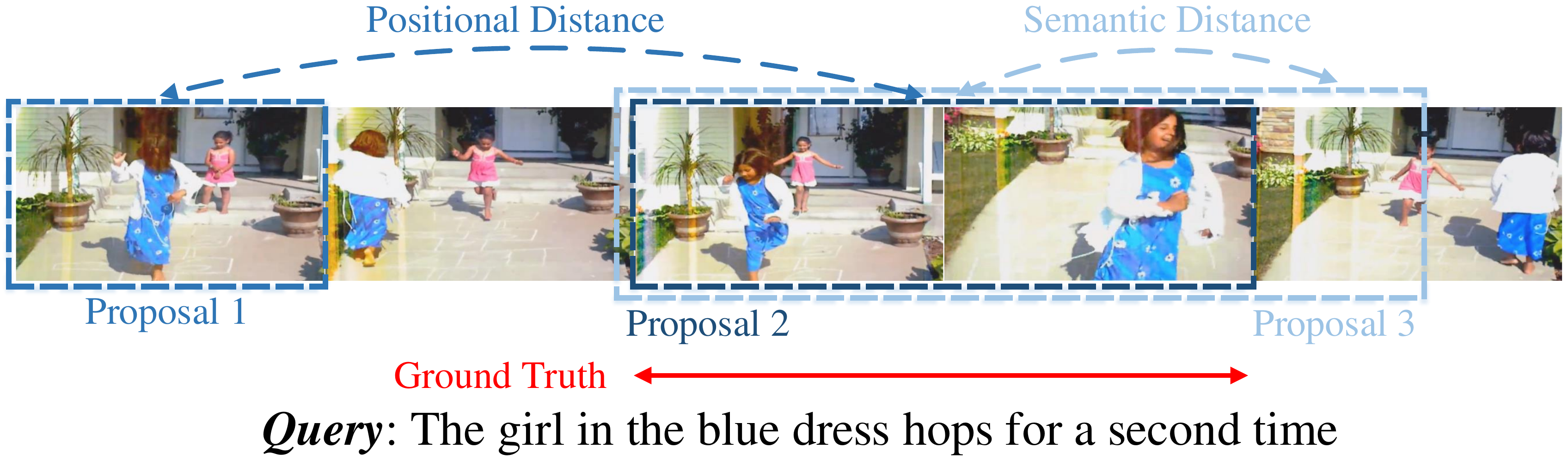}
\caption{To distinguish above three proposals, both positional and semantic relations among proposals needs to be considered.}
\label{fig:graph}
\vspace{-8pt}
\end{figure}

\subsection{Proposal Consolidation}
So far, we have generated a certain number of proposals that are significantly less than the pre-defined ones in existing \textit{top-down} framework, making the final scoring and ranking process much efficient. To further refine the proposal features for more accurate segment localization, we explicitly model higher-order interactions between the generated proposals to learn their relations. As shown in Figure \ref{fig:graph}, proposal 1 and proposal 2 contain same semantics of ``blue" and ``hops", we need to model their positional distance to distinguish them and refine their features for better understanding the phrase ``second time". Also, for the proposals (proposal 2 and 3) which are local neighbors, we have to learn their semantic distance to refine their representations. Therefore, in our APGN, we first encode each proposal feature with both positional embedding and frame-wise semantic features, and then define a graph convolutional network (GCN) over the proposals for proposal refinement.

\noindent \textbf{Proposal encoder.} For each proposal tuple $(t, l_{s}^{t}, l_{e}^{t})$, we represent its segment boundary as $(t-l_{s}^t, t+l_{e}^t)$. Before aggregating the features of its contained frames within this segment boundary, we first concatenate a position embedding $\bm{emb}^{pos}_t$ to each frame-wise feature $\widetilde{\bm{v}}_t$, in order to inject position information on frame $t$ as follows:
\begin{equation}
    \widetilde{\bm{v}}_t' = [\widetilde{\bm{v}}_t;\bm{emb}^{pos}_t] \in \mathbb{R}^{1\times (D+d)},
\end{equation}
where $\bm{emb}^{pos}_t$ denotes the position embedding of the $t$-th position, and $d$ is the dimension of $\bm{emb}^{pos}_t$. We follow \cite{vaswani2017attention} and use the sine and cosine functions of different frequencies to compose position embeddings:
\begin{equation}
    \bm{emb}^{pos}_t[2j]=\text{sin}(\frac{t}{10000^{2j/d}}),
\end{equation}
\begin{equation}
    \bm{emb}^{pos}_t[2j+1]=\text{cos}(\frac{t}{10000^{2j/d}}),
\end{equation}
where $2j$ and $2j+1$ are the even and odd indices of the position embedding. In this way, each dimension of the positional encoding corresponds to a sinusoid, allowing the model to easily learn to attend to absolute positions. Given the frame features $\{\widetilde{\bm{v}}_t'\}_{t=1}^{T_{fore}}$ and a proposal segment $(t-l_{s}^t, t+l_{e}^t)$, we encode the vector feature $\bm{p}_t$ of $t$-th proposal by aggregating the features of the contained frames in the segment as:
\begin{equation}
    \bm{p}_t = \text{MLP}_2(\text{Pool}(\text{MLP}_1([\widetilde{\bm{v}}_{\lceil t-l_{s}^t\rceil}',...,\widetilde{\bm{v}}_{\lfloor t+l_{e}^t\rfloor}']))),
\end{equation}
where each MLP has two FC layers, $\text{Pool}(\cdot)$ denotes the max-pooling. The frames from each proposal are independently processed by $\text{MLP}_1$ before being pooled (channel-wise) to a single feature vector and passed to $\text{MLP}_2$ where information from different frames are further combined. Thus, we can represent the encoded proposal feature as $\bm{p}_t \in \mathbb{R}^{1\times(D+d)}$.

\noindent \textbf{Proposal graph.} We construct a graph over the proposal features $\{\bm{p}_t\}_{t=1}^{T_{fore}}$, where each node of the graph is a proposal associated with both positions and semantic features. We full connect all node pairs, and define relations between each proposal-pair $(\bm{p}_{t},\bm{p}_{t'})$ for edge convolution \cite{2018Dynamic} as:
\begin{equation}
    \bm{e}_{t,t'} = \text{Relu}(\bm{p}_t\bm{\theta}_1 + (\bm{p}_{t'}-\bm{p}_t)\bm{\theta}_2),
\end{equation}
where $\bm{\theta_1}$ and $\bm{\theta_2}$ are learnable parameters. We update each proposal feature $\bm{p}_t$ to $\widehat{\bm{p}}_t$ as follow:
\begin{equation}
    \widehat{\bm{p}}_t = \text{MaxPool}(\bm{e}_{t}), \ \ \bm{e}_{t}=\{\bm{e}_{t,t'}\}_{t'=1}^{T_{fore}}.
\end{equation}
This GCN module consists of $k$ stacked graph convolutional layers. After the above proposal consolidation with graph, we are able to learn the refined proposal features.

\subsection{Localization Head}
After proposal consolidation, we feed the refined features $\widehat{\bm{P}}=\{\widehat{\bm{p}}_t\}_{t=1}^{T_{fore}}$ into two separate heads to predict their confidence scores and boundary offsets for proposal ranking and refinement. Specifically, we employ two MLPs on each feature $\widehat{\bm{p}}_t$ as:
\begin{equation}
    r_t = \text{Sigmoid}(\text{MLP}_3(\widehat{\bm{p}}_t)),
\end{equation}
\begin{equation}
    (\delta_{s}^t, \delta_{e}^t) = \text{MLP}_4(\widehat{\bm{p}}_t),
\end{equation}
where $r_t \in (0,1)$ is the confidence score, and $(\delta_{s}^t, \delta_{e}^t)$ is the offsets. Therefore, the final predicted segment of proposal $t$ can be represented as $(t-l_{s}^t+\delta_{s}^t, t+l_{e}^t+\delta_{e}^t)$. To learn the confidence scoring rule, we first compute the IoU score $o_t$ between each proposal segment with the ground-truth $(\tau_s,\tau_e)$, then we adopt the alignment loss function as below:
\begin{equation}
    \mathcal{L}_{align} = -\frac{1}{T_{fore}}\sum_{t=1}^{T_{fore}}o_t\text{log}(r_t)+(1-o_t)\text{log}(1-r_t).
\end{equation}
Given the ground-truth boundary offsets $(\hat{\delta}_{s}^t, \hat{\delta}_{e}^t)$ of proposal $t$, we also fine-tune its offsets by a boundary loss as:
\vspace{-10pt}
\begin{equation}
    \mathcal{L}_{b} = \frac{1}{T_{fore}}\sum_{t=1}^{T_{fore}}\text{SL}_1(\hat{\delta}_{s}^t-\delta_{s}^t)+\text{SL}_1(\hat{\delta}_{e}^t-\delta_{e}^t),
\end{equation}
where $\text{SL}_1(\cdot)$ denotes the smooth L1 loss function.

At last, our APGN model is trained end-to-end from scratch using the multi-task loss :
\begin{equation}
    \mathcal{L} = \lambda_1 \cdot \mathcal{L}_{class}+\lambda_2 \cdot \mathcal{L}_{reg}+ \lambda_3 \cdot \mathcal{L}_{align}+ \lambda_4 \cdot \mathcal{L}_{b}.
\end{equation}

\section{Experiments}
\subsection{Datasets and Evaluation}
\noindent \textbf{ActivityNet Captions.} It is a large dataset \cite{krishna2017dense} which contains 20k videos with 100k language descriptions. 
This dataset pays attention to more complicated human activities in daily life. 
Following public split, we use 37,417, 17,505, and 17,031 sentence-video pairs for training, validation, and testing, respectively.

\noindent \textbf{TACoS.} This dataset \cite{regneri2013grounding} collects 127 long videos, which are mainly about cooking scenarios, thus lacking the diversity. We use the same split as \cite{gao2017tall}, which has 10146, 4589 and 4083 sentence-video pairs for training, validation, and testing, respectively.

\noindent \textbf{Charades-STA.} \cite{gao2017tall} consists of 9,848 videos of daily life indoors activities. There are 12,408 sentence-video pairs for training and 3,720 pairs for testing.

\noindent \textbf{Evaluation Metric.} Following \cite{zhang2019cross,zeng2020dense}, we adopt “R@$n$, IoU=$m$” as our evaluation metrics, which is defined as the percentage of at least one of top-$n$ selected moments having IoU larger than $m$.

\subsection{Implementation Details}
Following \cite{zhang2019learning,zeng2020dense}, for video input, we apply a pre-trained C3D network for all three datasets to obtain embedded features.
We also extract the I3D \cite{carreira2017quo} and VGG \cite{simonyan2014very} features on Charades-STA. After that, we apply PCA to reduce their feature dimension to 500 for decreasing the model parameters. We set the length of video to 200 for ActivityNet Caption and TACoS, 64 for Charades-STA. For sentence input, we utilize Glove model to embed each word to 300 dimension features. The dimension $D$ is set to 512, $d$ is set to 256. The number of graph layer is $k=2$. We set the batchsize as 64.
We train our model with an Adam optimizer for 100 epochs. The initial learning rate is set to 0.0001 and it is divided by 10 when the loss arrives on plateaus. $\lambda_1, \lambda_2, \lambda_3, \lambda_4$ in the loss function are 0.1, 1, 1, 1 and decided by the weight magnitude.

\subsection{Performance Comparison}
\noindent \textbf{Compared methods.}
We compare our proposed APGN with state-of-the-art methods. We group them into: (1) \textit{top-down} methods: TGN \cite{chen2018temporally}, CTRL \cite{gao2017tall}, QSPN \cite{xu2019multilevel}, CBP \cite{wang2019temporally}, SCDM \cite{yuan2019semantic}, CMIN \cite{zhang2019cross}, and 2DTAN \cite{zhang2019learning}. 
(2) \textit{bottom-up} methods: GDP \cite{chenrethinking}, LGI \cite{mun2020LGI}, VSLNet \cite{zhang2020span}, DRN \cite{zeng2020dense}. 

\begin{table}[t!]
    \small
    \centering
    \setlength{\tabcolsep}{1.2mm}{
    \begin{tabular}{c|c|cccc}
    \toprule
    \multirow{2}*{Method} & \multirow{2}*{Feature} & R@1, & R@1, & R@5, & R@5 \\ 
    ~ & ~ & IoU=0.5 & IoU=0.7 & IoU=0.5 & IoU=0.7  \\ \midrule 
    TGN & C3D & 28.47 & - & 43.33 & - \\
    CTRL & C3D & 29.01 & 10.34 & 59.17 & 37.54  \\
    QSPN & C3D & 33.26 & 13.43 & 62.39 & 40.78  \\
    CBP & C3D & 35.76 & 17.80 & 65.89 & 46.20  \\
    SCDM & C3D & 36.75 & 19.86 & 64.99 & 41.53  \\
    GDP & C3D & 39.27 & - & - & -  \\
    LGI & C3D & 41.51 & 23.07 & - & - \\
    VSLNet & C3D & 43.22 & 26.16 & - & -  \\
    CMIN & C3D & 43.40 & 23.88 & 67.95 & 50.73 \\
    DRN & C3D & 45.45 & 24.36 & 77.97 & 50.30 \\ 
    2DTAN & C3D & 44.51 & 26.54 & 77.13 & 61.96 \\   \midrule
    \textbf{APGN} & C3D & \textbf{48.92} & \textbf{28.64} & \textbf{78.87} & \textbf{63.19} \\ \bottomrule
    \end{tabular}}
    \caption{Performance compared with the state-of-the-art TSLV models on ActivityNet Captions dataset.}
    \vspace{-10pt}
    \label{tab:compare}
\end{table}

\begin{table}[t!]
    \small
    \centering
    \setlength{\tabcolsep}{1.2mm}{
    \begin{tabular}{c|c|cccc}
    \toprule
    \multirow{2}*{Method} & \multirow{2}*{Feature} & R@1, & R@1, & R@5, & R@5,  \\ 
    ~ & ~ & IoU=0.3 & IoU=0.5 & IoU=0.3 & IoU=0.5 \\ \midrule 
    TGN & C3D & 21.77 & 18.90 & 39.06 & 31.02 \\
    CTRL & C3D & 18.32 & 13.30 & 36.69 & 25.42 \\
    QSPN & C3D & 20.15 & 15.23 & 36.72 & 25.30  \\
    CBP & C3D & 27.31 & 24.79 & 43.64 & 37.40  \\
    SCDM & C3D & 26.11 & 21.17 & 40.16 & 32.18 \\
    GDP & C3D & 24.14 & - & - & - \\
    VSLNet & C3D & 29.61 & 24.27 & - & - \\
    CMIN & C3D & 24.64 & 18.05 & 38.46 & 27.02 \\
    DRN & C3D & - & 23.17 & - & 33.36 \\ 
    2DTAN & C3D & 37.29 & 25.32 & 57.81 & 45.04  \\\midrule
    \textbf{APGN} & C3D & \textbf{40.47} & \textbf{27.86} & \textbf{59.98} & \textbf{47.12} \\ \bottomrule
    \end{tabular}}
    \caption{Performance compared with the state-of-the-art TSLV models on TACoS datasets.}
    \label{tab:compare2}
\end{table}

\begin{table}[t!]
    \small
    \centering
    \setlength{\tabcolsep}{1.2mm}{
    \begin{tabular}{c|c|cccc}
    \toprule 
    \multirow{2}*{Method} & \multirow{2}*{Feature} & R@1, & R@1, & R@5, & R@5, \\ 
    ~ & ~ & IoU=0.5 & IoU=0.7 & IoU=0.5 & IoU=0.7 \\ \midrule 
    2DTAN & VGG & 39.81 & 23.25 & 79.33 & 51.15 \\
    \textbf{APGN} & VGG & \textbf{44.23} & \textbf{25.64} & \textbf{89.51} & \textbf{57.87} \\ \midrule
    CTRL & C3D & 23.63 & 8.89 & 58.92 & 29.57 \\
    QSPN & C3D & 35.60 & 15.80 & 79.40 & 45.40 \\
    CBP & C3D & 36.80 & 18.87 & 70.94 & 50.19 \\
    GDP & C3D & 39.47 & 18.49 & - & - \\
    \textbf{APGN} & C3D & \textbf{48.20} & \textbf{29.37} & \textbf{89.05} & \textbf{58.49} \\ \midrule
    DRN & I3D & 53.09 & 31.75 & 89.06 & 60.05 \\ 
    SCDM & I3D & 54.44 & 33.43 & 74.43 & 58.08 \\
    LGI & I3D & 59.46 & 35.48 & - & - \\
    \textbf{APGN} & I3D & \textbf{62.58} & \textbf{38.86} & \textbf{91.24} & \textbf{62.11} \\ \bottomrule
    \end{tabular}}
    \caption{Performance compared with the state-of-the-art TSLV models on Charades-STA datasets.}
    \label{tab:compare3}
    \vspace{-10pt}
\end{table}

\noindent \textbf{Quantitative comparison.} As shown in Table \ref{tab:compare}, \ref{tab:compare2} and \ref{tab:compare3}, our APGN outperforms all the existing methods by a large margin. Specifically, on ActivityNet Caption dataset, compared to the previous best \textit{top-down} method 2DTAN, we do not rely on large numbers of pre-defined and outperform it by 4.41\%, 2.10\%, 1.74\%, 1.23\% in all metrics, respectively. Compared to the previous best \textit{bottom-up} method DRN, our APGN brings significant improvement of 4.28\% and 12.89\% in the strict ``R@1, IoU=0.7” and ``R@5, IoU=0.7” metrics, respectively.
Although TACoS suffers from similar kitchen background and cooking objects among the videos, it is worth noting that our APGN still achieves significant improvements.
On Charades-STA dataset, for fair comparisons with other methods, we perform experiments with same features (i.e., VGG, C3D, and I3D) reported in their papers. It shows that our APGN reaches the highest results over all evaluation metrics. 

\noindent \textbf{Comparison on efficiency.} We compare the efficiency of our APGN with previous methods 
on a single Nvidia TITAN XP GPU on the TACoS dataset. As shown in Table \ref{tab:efficient}, it can be observed that we achieve much faster processing speeds and relatively less learnable parameters. The reason mainly owes to two folds: First, APGN generates proposals without processing overlapped sliding windows as CTRL, and generates less proposals than pre-defined methods such as 2DTAN and CMIN, thus is more efficient; Second, APGN does not apply many convolution layers like 2DTAN or multi-level feature fusion modules as DRN for cross-modal interaction, thus has less parameters.

\begin{table}[t!]
    \small
    \centering
    \setlength{\tabcolsep}{0.7mm}{
    \begin{tabular}{c|ccccccc}
    \toprule 
    ~ & ACRN & CTRL & TGN & 2DTAN & CMIN & DRN & \textbf{APGN} \\ \midrule
    VPS $\uparrow$ & 0.23 & 0.45 & 1.09 & 1.75 & 81.29 & 133.38 & \textbf{146.67} \\ \midrule
    Para. $\downarrow$ & 128 & \textbf{22} & 166 & 363 & 78 & 214 & 91 \\ \bottomrule
    \end{tabular}}
    \caption{Efficiency comparison in terms of video per second (VPS) and parameters (Para.), where our method APGN is much efficient.}
    \label{tab:efficient}
\end{table}

\begin{table}[t!]
    \small
    \centering
    \setlength{\tabcolsep}{1.6mm}{
    \begin{tabular}{c|cccc|cc}
    \toprule 
    \multirow{2}*{Model} & \multirow{2}*{class.} & \multirow{2}*{reg.} & \multirow{2}*{p.e.} & \multirow{2}*{graph} & R@1, & R@1, \\ 
    ~ & ~ & ~ & ~ & ~ & IoU=0.5 & IoU=0.7 \\ \midrule 
    \ding{172} &$\times$ & $\times$ & $\times$ & $\times$ & 39.16 & 19.68 \\ \midrule
    \ding{173} & $\checkmark$ & $\times$ & $\times$ & $\times$ & 40.84 & 21.30 \\
    \ding{174} & $\checkmark$ & $\checkmark$  & $\times$ & $\times$ & 42.77 & 23.52 \\
    \ding{175} & $\checkmark$ & $\checkmark$ & $\checkmark$ & $\times$ & 43.95 & 24.66 \\
    \ding{176} & $\checkmark$ & $\checkmark$ & $\times$ & $\checkmark$ & 45.81 & 26.34 \\
    \ding{177} & $\checkmark$ & $\checkmark$ & $\checkmark$ & $\checkmark$ & \textbf{48.92} & \textbf{28.64} \\ \bottomrule
    \end{tabular}}
    \caption{Main ablation studies on ActivityNet Caption dataset, where `class.' and `reg.' denotes the classification and regression modules (Sec. 3.3), `p.e' denotes the proposal encoder (Sec. 3.4), `graph' denotes the proposal graph (Sec. 3.4).}
    \label{tab:ablation1}
    \vspace{-10pt}
\end{table}

\subsection{Ablation Study}
\noindent \textbf{Main ablation.} As shown in Table \ref{tab:ablation1}, we verify the contribution of each part in our model. Starting from the backbone model (Figure \ref{fig:pipeline} (a)), we first implement the baseline model \ding{172} by directly adding the \textit{top-down} localization head ((Figure \ref{fig:pipeline} (d))). In this model, we adopt pre-defined proposals as \cite{zhang2019cross}.
After adding the binary classification module in \ding{173}, we can find that classification module effectively filters out redundant pre-defined proposals on large number of background frames. When further applying adaptive proposal generation as \ding{174}, the generated proposals perform better than the pre-defined one \ding{173}. Note that, in \ding{174}, we directly encode proposal-wise features by max-pooling, and the classification module also makes the contribution for filtering out the negative generated proposals.
To capture more fine-grained semantics for proposal refinement, we introduce a proposal encoder (model \ding{175}) for discriminative feature aggregation and a proposal graph (model \ding{176}) for proposal-wise feature interaction. Although each of them can only bring about 1-3\% improvement, the performance increases significantly when utilizing both of them (model \ding{177}).

\noindent \textbf{Investigation on the video/query encoder.} To investigate whether a Transformer \cite{vaswani2017attention} can boost our APGN, we replace the GRU in video/query encoder with a simple Transformer and find some improvements. However, it brings larger model parameters and lower speed.

\noindent \textbf{Effect of unbalanced loss.} In the binary classification module, 
we formulate the typical loss function into a balanced one. As shown in Table \ref{tab:ablation2}, the model w/ balanced loss has great improvement (2.04\%, 1.51\%) compared to the w/o variant, which demonstrates that it is important to consider the unbalanced distribution in the classification process.

\noindent \textbf{Investigation on proposal encoder.} In proposal encoder, we discard the positional embedding as w/o position, and also replace the max-pooling with the mean-pooling as w/ mean pooling. From the Table \ref{tab:ablation3}, we can observe that positional embedding helps to learn the temporal distance (boost 2.46\%, 1.95\%), and the max-pooling can aggregate more discriminative features (boost 1.49\%, 0.78\%) than the mean-pooling.

\noindent \textbf{Investigation on proposal graph.} In the table \ref{tab:ablation4}, we also give the analysis on the proposal graph. Compared to w/ edge convolution model \cite{2018Dynamic}, w/ edge attention directly utilizes co-attention \cite{lu2016hierarchical} to compute the similarity of each node-pair and updates them by a weighted summation strategy, which performs worse than the former one.


\begin{table}[t!]
    \small
    \centering
    \setlength{\tabcolsep}{1.2mm}{
    \begin{tabular}{c|c|c|cc}
    \toprule
    \multirow{2}*{Components} & \multirow{2}*{VPS $\uparrow$} & \multirow{2}*{Para. $\downarrow$} & R@1, & R@1, \\ 
    ~ & ~ & ~ & IoU=0.5 & IoU=0.7 \\ \midrule
    w/. GRU & \textbf{146.67} & \textbf{91} & 48.92 & 28.64 \\
    w/. Transformer & 129.38 & 138 & \textbf{50.11} & \textbf{29.43} \\ \bottomrule
    \end{tabular}}
    \caption{Investigation on video and query encoders on ActivityNet Caption dataset.}
    \label{tab:ablation_add}
\end{table}

\begin{table}[t!]
    \small
    \centering
    \setlength{\tabcolsep}{1.2mm}{
    \begin{tabular}{c|c|cc}
    \toprule
    \multirow{2}*{Components} & \multirow{2}*{Module} & R@1, & R@1, \\ 
    ~ & ~ & IoU=0.5 & IoU=0.7 \\ \midrule
    \multirow{2}*{\tabincell{c}{binary\\classification}} & w/o balanced loss & 46.88 & 27.13 \\
    ~ & w/ balanced loss & \textbf{48.92} & \textbf{28.64} \\ \bottomrule
    \end{tabular}}
    \caption{Investigation on binary classification on ActivityNet Caption dataset.}
    \label{tab:ablation2}
\end{table}

\begin{table}[t!]
    \small
    \centering
    \setlength{\tabcolsep}{1.2mm}{
    \begin{tabular}{c|c|cc}
    \toprule
    \multirow{2}*{Components} & \multirow{2}*{Module} & R@1, & R@1, \\ 
    ~ & ~ & IoU=0.5 & IoU=0.7 \\ \midrule
    \multirow{4}*{\tabincell{c}{proposal\\encoder}} & w/o position & 46.46 & 26.69 \\
    ~ & w/ position & \textbf{48.92} & \textbf{28.64} \\ \cline{2-4}
     ~ & w/ mean pooling & 47.41 & 27.86 \\
    ~ & w/ max pooling & \textbf{48.92} & \textbf{28.64} \\ \bottomrule
    \end{tabular}}
    \caption{Investigation on proposal encoder on ActivityNet Caption dataset.}
    \label{tab:ablation3}
    \vspace{-10pt}
\end{table}

\begin{table}[t!]
    \small
    \centering
    \setlength{\tabcolsep}{1.2mm}{
    \begin{tabular}{c|c|cc}
    \toprule
    \multirow{2}*{Components} & \multirow{2}*{Module} & R@1, & R@1, \\ 
    ~ & ~ & IoU=0.5 & IoU=0.7 \\ \midrule
    \multirow{2}*{\tabincell{c}{proposal\\graph}} & w/ edge attention & 46.63 & 26.90 \\
    ~ & w/ edge convolution & \textbf{48.92} & \textbf{28.64} \\ \midrule
    \multirow{3}*{\tabincell{c}{graph\\layer}} & 1 layer & 47.60 & 27.57 \\
    ~ & 2 layers & \textbf{48.92} & \textbf{28.64} \\
    ~ & 3 layers & 48.83 & 28.39 \\ \bottomrule
    \end{tabular}}
    \caption{Investigation on proposal graph on ActivityNet Caption dataset.}
    \label{tab:ablation4}
\end{table}

\begin{table}[t!]
    \small
    \centering
    \setlength{\tabcolsep}{1.2mm}{
    \begin{tabular}{c|c|cc}
    \toprule
    \multirow{2}*{Methods} & Localization & R@1, & R@1, \\ 
    ~ & Type & IoU=0.5 & IoU=0.7 \\ \midrule
    \multirow{2}*{SCDM} & \textit{top-down} & 36.75 & 19.86  \\
    ~ & ours & \textbf{43.86} & \textbf{26.42} \\ \midrule
    \multirow{2}*{CMIN} & \textit{top-down} & 43.40 & 23.88  \\
    ~ & ours & \textbf{50.33} & \textbf{29.75} \\ \midrule
    \multirow{2}*{LGI} & \textit{bottom-up} & 41.51 & 23.07  \\
    ~ & ours & \textbf{49.20} & \textbf{30.64}  \\ \midrule
    \multirow{2}*{DRN} & \textit{bottom-up} & 45.45 & 24.36  \\
    ~ & ours & \textbf{53.72} & \textbf{31.01} \\ \bottomrule
    \end{tabular}}
    \caption{Our proposed adaptive proposal generation can serve as a ``plug-and-play" module for existing methods. The experiments are conducted on the ActivityNet Captions dataset.}
    \label{tab:ablation5}
    \vspace{-10pt}
\end{table}

\noindent \textbf{Number of graph layer.} 
As shown in Table \ref{tab:ablation4}, the model achieves the best result with 2 graph layers, and the performance will drop when the number of layers grows up. We give the analysis is that more graph layers will result in over-smoothing problem \cite{li2018deeper} since the propagation between the nodes will be accumulated.

\noindent \textbf{Plug-and-play.}
Our proposed adaptive proposal generation can serve as a plug-and-play for existing methods. As shown in Table \ref{tab:ablation5}, for \textit{top-down} methods, we maintain their feature encoders and video-query interaction, and add the proposal generation and proposal consolidation before the localization heads. For \textit{bottom-up} methods, we first replace their regression heads with our proposal generation process and then add the proposal consolidation process. It shows that our proposal generation and proposal consolidation can bring large improvement on both two types of methods.

\subsection{Qualitative Results}
\begin{figure}[t!]
\centering
\includegraphics[width=0.48\textwidth]{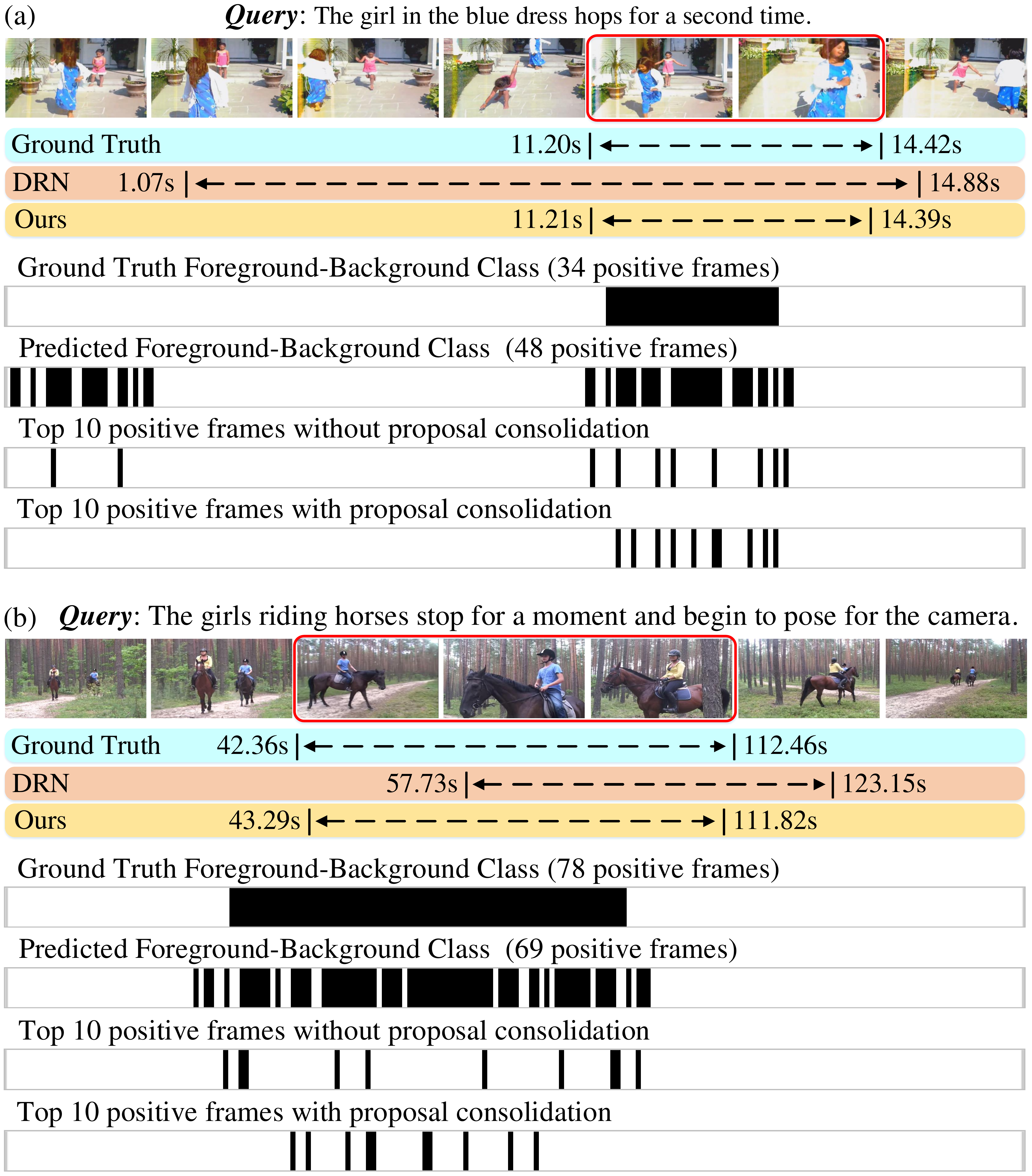}
\caption{Typical examples of the localization results on the ActivityNet Caption dataset.}
\label{fig:result}
\vspace{-8pt}
\end{figure}

To qualitatively validate the effectiveness of our APGN, we display two typical examples in Figure \ref{fig:result}. It is challenging to accurately localize the semantic ``for a second time" in the first video, because there are two separate segments corresponding to the same object ``girl in the blue dress" performing the same activity ``hops". For comparison, previous method DRN fails to understand the meaning of phrase ``second time", and ground both two segment parts. By contrast, our method has a strong ability to distinguish these two segments in temporal dimension thanks to the positional embedding in the developed proposal graph, thus achieves more accurate localization results.
Furthermore, we also display the foreground/background class of each frame in this video. With the help of the proposal consolidation module, the segment proposals of ``first time" are filtered out, and all the final ranked top 10 positive frames fall in the target segment.

\section{Conclusion}
In this paper, we introduce APGN, a new method for temporal sentence localization in videos. Our core idea is to adaptively generates discriminative proposals and achieve both effective and efficient localization.
That is, we first introduce binary classification before the boundary regression to distinguish the background frames, which helps to filter out the corresponding noisy responses. Then, the regressed boundaries on the predicted foreground frames are taken as segment proposals, which decreases a large number of poor quality proposals compared to the pre-defined ones in \textit{top-down} framework. We further learn higher-level feature interactions between the generated proposals for refinement via a graph convolutional network.
Our framework achieves state-of-the-art performance on three challenging benchmarks, demonstrating the effectiveness of our proposed APGN.

\section{Acknowledgments}

This work was supported in part by the National Key Research and Development Program of China under No. 2018YFB1404102, and the National Natural Science Foundation of China under No. 61972448.


\bibliography{anthology}
\bibliographystyle{acl_natbib}

\end{document}